\documentclass[letterpaper]{article} % DO NOT CHANGE THIS
\usepackage{aaai25}  % DO NOT CHANGE THIS
\usepackage{times}  % DO NOT CHANGE THIS
\usepackage{helvet}  % DO NOT CHANGE THIS
\usepackage{courier}  % DO NOT CHANGE THIS
\usepackage[hyphens]{url}  % DO NOT CHANGE THIS
\usepackage{graphicx} % DO NOT CHANGE THIS
\usepackage{natbib}  % DO NOT CHANGE THIS AND DO NOT ADD ANY OPTIONS TO IT
\usepackage{caption} % DO NOT CHANGE THIS AND DO NOT ADD ANY OPTIONS TO IT
\usepackage{algorithm}
\usepackage{algorithmic}
\usepackage{booktabs}
\usepackage{multirow}

\usepackage{newfloat}
\usepackage{listings}
\DeclareCaptionStyle{ruled}{labelfont=normalfont,labelsep=colon,strut=off} % DO NOT CHANGE THIS
\floatstyle{ruled}
\newfloat{listing}{tb}{lst}{}
\floatname{listing}{Listing}
\title{Enhancing Chain of Thought Prompting in Large Language Models \\ via Reasoning Patterns}
\author{
    Yufeng Zhang\textsuperscript{\rm 1,2}, Xuepeng Wang\textsuperscript{\rm 1,3,}\thanks{Corresponding authors.}, Lingxiang Wu\textsuperscript{\rm 1,3,}\footnotemark[1], Jinqiao Wang\textsuperscript{\rm 1,2,3}
}
\affiliations{
    \textsuperscript{\rm 1}Institute of Automation, Chinese Academy of Sciences, Beijing, China\\
    \textsuperscript{\rm 2}School of Artificial Intelligence, University of Chinese Academy of Sciences, Beijing, China\\
    \textsuperscript{\rm 3}Wuhan AI Research, Wuhan, China\\
    \{yufeng.zhang, xuepeng.wang\}@ia.ac.cn,
    \{lingxiang.wu, jqwang\}@nlpr.ia.ac.cn
}

\usepackage{bibentry}
\begin{document}

\maketitle

\begin{abstract}
Chain of Thought (CoT) prompting can encourage language models to engage in multi-step logical reasoning. The quality of the provided demonstrations significantly influences the success of downstream inference tasks. Current unsupervised CoT methods primarily select examples based on the semantics of the questions, which can introduce noise and lack interpretability. In this paper, we propose leveraging reasoning patterns to enhance CoT prompting effectiveness. Reasoning patterns represent the process by which language models arrive at their final results. By utilizing prior knowledge and prompt-based methods from large models, we first construct task-specific pattern sets. We then select diverse demonstrations based on different reasoning patterns. This approach not only mitigates the impact of noise but also provides explicit interpretability to help us understand the mechanisms of CoT. Extensive experiments demonstrate that our method is more robust and consistently leads to improvements across various reasoning tasks.
\end{abstract}


\section{Introduction}

Large Language Models (LLMs) have demonstrated exceptional performance across a wide range of language tasks. In general question-answering tasks \cite{kwiatkowski2019natural}, LLMs hold a distinct advantage over other language models due to their robust writing capabilities. However, when it comes to more advanced tasks such as logical reasoning, mathematical computation, and symbolic reasoning, LLMs often fall short \cite{qiao2022reasoning,huang2022towards}.

\begin{figure}[t]
    \centering
    \includegraphics[width=0.43\textwidth]{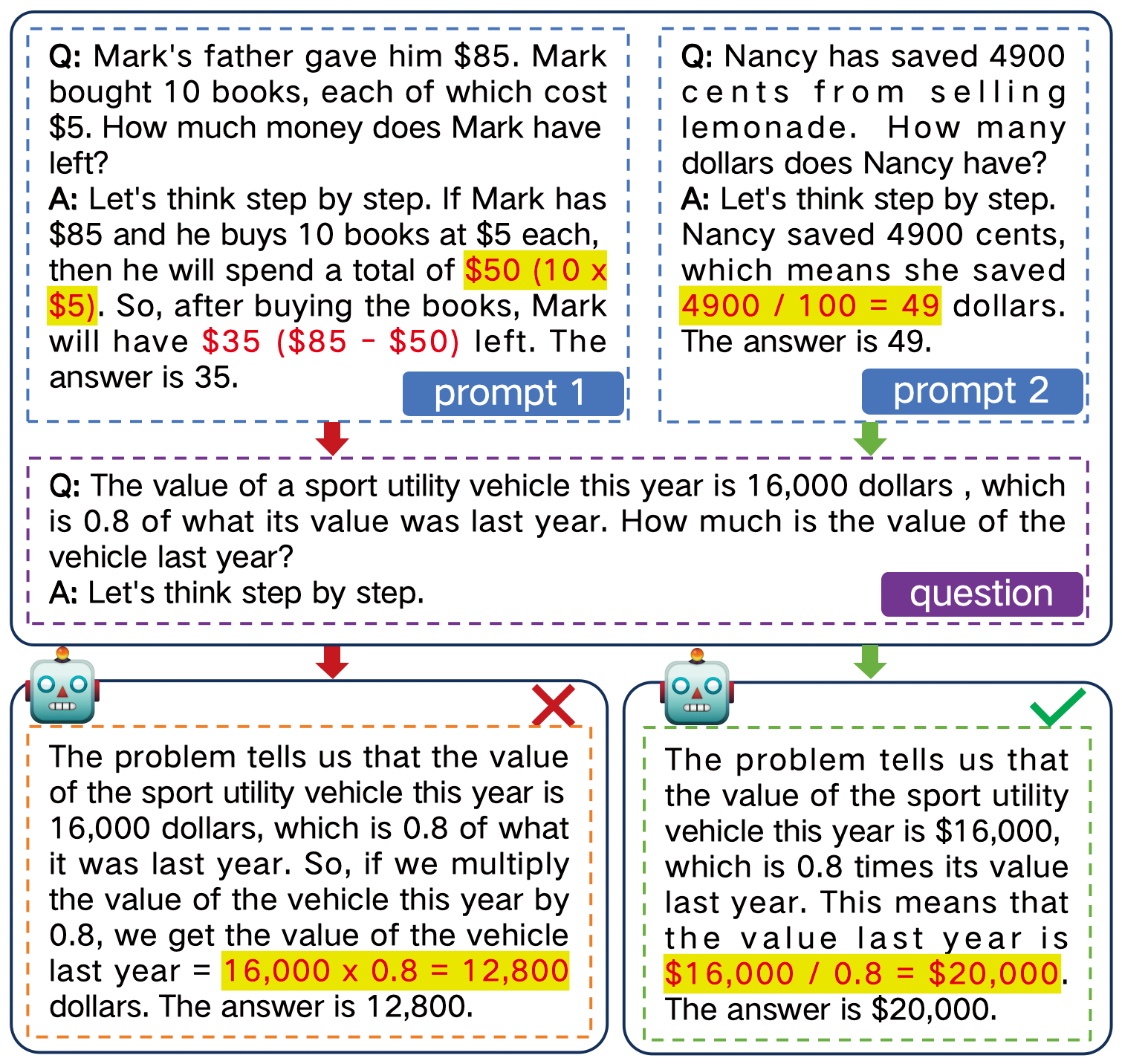}
    \caption{Example of the chain-of-thought prompting. The prompt influences how LLMs arrive at the final answer.}
    \label{fig:example}
\end{figure}

One effective approach to addressing these challenges is Chain of Thought (CoT) prompting \cite{wei2022chain}. By providing several demonstration examples that include a problem, intermediate reasoning steps, and an answer, CoT prompting serves as a contextual guide for downstream tasks. This approach encourages LLMs to generate multi-step logical reasoning, thereby maximizing the likelihood of producing more plausible answers. The advantage of this method lies in its simplicity and efficiency; unlike fine-tuning, it does not require extensive gradient updates or alter the model's inherent capabilities. Instead, it acts as an external augmentation of knowledge. For different reasoning tasks, we can route the model to the appropriate context, and then easily switch the demonstration sets to activate the relevant knowledge and abilities in the corresponding domain.

However, we argue that existing unsupervised CoT prompting methods have two major shortcomings. First, there remains a significant gap between the selected demonstration sets and the reasoning targets. Although extensive research \cite{zhang2023automatic,levy2022diverse,yang2023representative,shum-etal-2023-automatic} has explored ways to provide CoT demonstrations to enhance LLMs' reasoning capabilities, these methods largely rely on the semantic features of the problem or the answer. Such features introduce irrelevant noise on a global scale, which can obscure the logical information needed for reasoning. Consequently, the constructed demonstration sets do not effectively represent the domain-specific logical knowledge, and struggle to adequately trigger correct reasoning in LLMs. Second, some demonstration selection methods lack interpretability and scalability. These methods are primarily based on heuristic design \cite{wang2022rationale,zheng2023progressive} or leverage the model itself to generate additional demonstrations \cite{zhong2024achieving,yasunaga2024large}. The demonstration sets chosen through these means inherently lack clear explanations, making it challenging to assess their effectiveness or determine the direction for further optimization. This limitation can be particularly problematic in scenarios where interpretability is crucial.

To better select a demonstration subset for a reasoning task, we believe that considering the logical patterns of reasoning is essential. Inspired by the work of \cite{min2022rethinking} and \cite{madaan2023makes}, we observe that LLMs are more influenced by the templates and patterns in the context than by the correctness of the demonstrations themselves. Building on this insight, we investigate the selection of demonstrations based on \textbf{Reasoning Patterns}. This approach offers a dual benefit. First, it helps to eliminate bias introduced by irrelevant information, thereby reducing the gap between the demonstration set and the reasoning task. Second, it provides explicit interpretability, allowing us to gain a deeper understanding of how CoT prompting functions. This interpretability can also serve as a clue for attribution analysis and visualization.

In this work, we propose \textbf{Pattern-CoT}\footnote{ https://github.com/Magicat128/Pattern-CoT.}, a CoT demonstration selection method based on reasoning patterns. Unlike previous approaches that focus on overall semantics, our method targets finer-grained logical reasoning operations. For instance, in mathematical reasoning, addition and multiplication represent distinct operations, while multiple sequential operators may indicate more complex operational patterns, as shown in Figure \ref{fig:example}. Inspired by recent studies \cite{yang2023representative}, a diverse range of these patterns should be incorporated into CoT. Specifically, for a given reasoning task, we first obtain a set of seed demonstrations with rationale (intermediate reasoning steps). These examples can be sourced from the training set or generated using a zero-shot approach. We then obtain specific operation tokens tailored to different task types, which help us extract reasoning patterns from the rationales. Here, we incorporate prior knowledge and guide the LLMs in generating these operation tokens. Based on the extracted reasoning patterns, we apply clustering techniques to merge similar patterns and design metrics to automatically assess the number of demonstration categories. Finally, we select representative demonstrations from each category to enrich the diversity and construct context prompts for LLMs. Notably, by incorporating task-specific knowledge, our method improves interpretability and facilitates further scalability.

Our contributions can be summarized as follows:
\begin{itemize}
    \item We introduce the use of diverse reasoning patterns to enhance CoT prompting and design a demonstration selection method to reduce the gap between the demonstration set and the task.
    \item Our method strengthens the interpretability of CoT in unsupervised scenarios, and can be utilized for further attribution analysis.
    \item Extensive experiments demonstrate that our method consistently enhances performance across multiple reasoning tasks and various models.
\end{itemize}

\section{Related Work}

\subsection{Chain-of-Thought Prompting}
Large language models have demonstrated significant ability in comprehending context and responding to prompts \cite{brown2020language,ouyang2022training}. Recent studies highlight that LLMs can achieve improved task completion without fine-tuning, particularly on reasoning tasks, when provided with few-shot demonstrations \cite{wei2022chain}. For instance, when presented with an example like \textit{Q: Mary has 9 yellow marbles. John has 3 yellow marbles. How many yellow marbles do they have in all? A: They have 9 + 3 = 12 yellow marbles. The answer is 12}, LLMs are expected to emulate such a format, deconstruct the question, engage in multi-step reasoning, and refrain from generating random answers in subsequent tasks. This process is commonly referred to as chain-of-thought prompting or in-context learning \cite{wei2022emergent,xie2022an}. However, implementing this practice often involves the manual design of prompts at a labour cost. Consequently, researchers are exploring more efficient example selection strategies to streamline this process.

\subsection{Demonstration Selection and Refinement}
Several CoT studies are directed towards automating the generation of demonstrations, such as retrieval-based \cite{rubin-etal-2022-learning}, zero-shot \cite{kojima2022large}, clustering-based \cite{zhang2023automatic}, and self-prompt \cite{shao2023synthetic,yasunaga2024large}. However, many of these approaches encounter challenges in achieving performance comparable to Manual-CoT, primarily due to the absence of supervision in example selection. In another branch of research, efforts are focused on enhancing the quality of CoT demonstrations. They incorporate elements such as knowledge-infusion \cite{zhao-etal-2023-verify, weng2023large,li2024cok}, self-consistency \cite{wang2023self}, complexity-based \cite{fu2022complexity}, contrastive-based \cite{chia2023contrastive}, and progressive-hint \cite{zheng2023progressive}. The primary goal of these strategies is to ensure that LLMs adhere to the correct prompt and avoid being misled.

\begin{figure*}[t]
    \centering
    \includegraphics[width=0.85\textwidth]{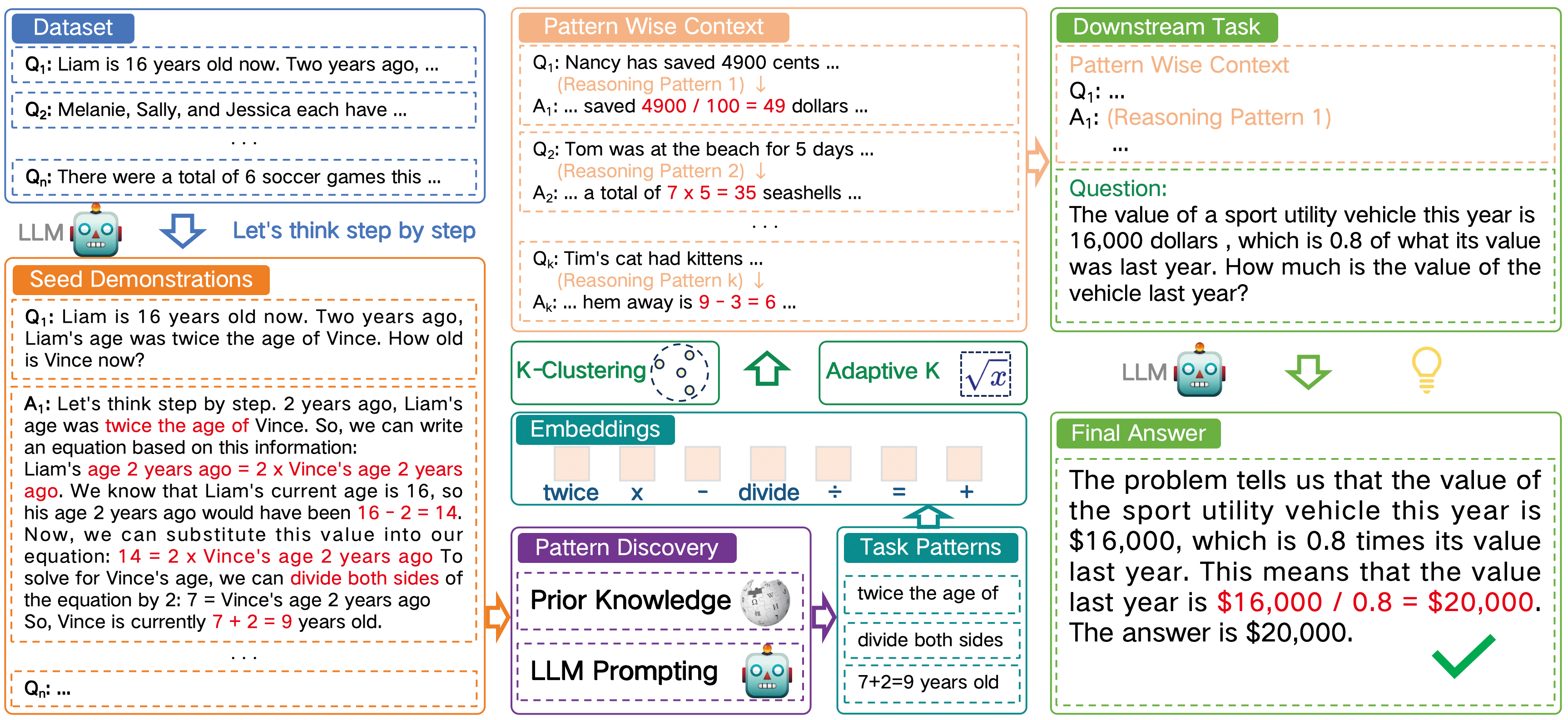}
    \caption{Illustration of our proposed framework. We first extract different patterns from the original rationales. Then clustering is used to produce a group of demonstrations. This enables LLMs to perceive diverse reasoning patterns and to select a proper solution path. It avoids LLMs being biased by monotonous reasoning mode.}
    \label{fig:main}
\end{figure*}

\subsection{Role of In-Context Patterns}
To understand the underlying mechanism of ICL, \cite{min2022rethinking} and \cite{madaan2023makes} employ counterfactual prompting methods. These methods involve substituting question-answer mapping, token distributions, answer patterns, and many other factors. Their findings consistently show that the correctness of examples is not the most crucial factor, but rather the distribution or pattern (e.g. equations, templates, sentence structure) of the examples. In this paper, we continue to uncover the power of CoT patterns and show how they can improve the reasoning process. 

\section{Methodology}
We now explore the impact of diverse demonstration reasoning patterns on chain-of-thought prompting. According to \cite{min2022rethinking}, the precision of demonstrations is not crucial when LLMs engage in ICL. Even if all the demonstrations provided are incorrect, it would only marginally impede performance. This aligns with the insight derived from Auto-CoT \cite{zhang2023automatic}: clustering zero-shot question-answer pairs without emphasizing accuracy can still yield valuable examples. Consequently, our focus shifts to a more nuanced factor - the underlying reasoning pattern that harbours more informative content \cite{madaan2023makes} - to evaluate its potential benefits for the CoT process. The entire process is summarized in Figure \ref{fig:main} and Algorithm \ref{alg:algorithm}.

\subsection{Seed Demonstration Collection}
For a given task $Q = \{q_1, q_2, ..., q_N\}$  with $N$ questions, we first need to obtain their rationales and answers $\{q_i, r_i, a_i\}$ that can be used as context for CoT prompting. For data from existing training sets, we can directly use the training data. However, in practical applications, complete training sets may not always be available. In such cases, we refer to methods like \cite{zhang2023automatic,shum2023automatic} and leverage the zero-shot \cite{kojima2022large} capabilities of LLMs to generate the corresponding rationales. It is important to note that we do not require the answers to be correct or labelled; our focus is on whether the generated rationales contain meaningful reasoning patterns.

\subsection{Pattern Discovery}
Based on the rationale set $Ra = \{r_1, r_2, ..., r_N\}$ that we have obtained, we next identify the reasoning operations $T$ associated with the task. For tasks with a relatively limited action space, we can define reasoning operations using prior knowledge, as these operations represent the fundamental units of reasoning tasks. For example, in arithmetic problems, we refer to a glossary of possible operators from sources like Wikipedia\footnote{The glossary of arithmetic operators refers to the Wikipedia: https://en.wikipedia.org/wiki/Glossary\_of\_mathematical\_symbols}, including basic arithmetic operations, square roots, comparison symbols, etc. For tasks with less clearly defined operations, we adapt definitions from arithmetic problems to guide LLMs in generating the corresponding reasoning operations. We design the prompt as: \textit{`Similar to operators used in arithmetic such as (+, -, *, /), which operators do you think best represent the [TASK]? Example of [TASK]: ...'}

For each rationale $r_i \in Ra$, we extract the reasoning operation tokens or phrases $t_j \in T$ to form its reasoning pattern:
\begin{equation}
    p_i = f(r_i, T) = \{t_{i1}, t_{i2}, ..., t_{ij}\}
\end{equation}
where $f$ denotes the function used to extract the reasoning path. In this context, $p_i$ represents how LLMs apply these operations step-by-step to reach the final result, and $t_{ij}$ can repeated.

\begin{algorithm}[htb]
    \caption{Pattern-CoT Demonstration Selection}
    \label{alg:algorithm}
    \begin{algorithmic}[1]
        \REQUIRE A set of task questions $Q$
        \ENSURE Demonstration list $d = [d_1, d_2, ..., d_k]$
        \STATE Acquire operation token set $T$ with LLMs prompting or domain knowledge based on $Q$
        \FOR {$q_i \in Q$}
            \STATE Generate rationale $r_i$ with Zero-Shot-CoT
            \STATE $p_i = [ ]$
            \FOR{each token $t_{ij} \in r_i$}
                \IF{$t_{ij} \in T$}
                    \STATE Update $p_i$ with $t_{ij}$
                \ENDIF
            \ENDFOR
            \STATE $\widetilde{p}_i = \mathrm{encode}(p_i)$
        \ENDFOR
        \STATE Select proper $k$
        \STATE Cluster all $[\widetilde{p}_1,\widetilde{p}_2,...,\widetilde{p}_i]$ into $k$ clusters
        \STATE Sample $d = [d_1, d_2, ..., d_k]$ from each cluster
        \RETURN $d$
    \end{algorithmic}
\end{algorithm}

\subsection{Pattern Wise Demonstration Selection}
Once we have identified the task-relevant patterns, we use them to select better demonstration sets. Following \cite{zhang2023automatic}, we cluster all the $p_i$ patterns while preserving diversity. Although $p_i$ is a simplified sequence of tokens, it still contains substantial semantic information that can be used to uncover underlying similarities. For instance, a sequence of addition operations is likely to be closer to a single addition operation than to a single multiplication operation. To leverage this, we use a language model to encode these patterns. We then apply the $k$-means clustering algorithm to generate $k$ clusters and sample from each cluster:
\begin{equation}
    \widetilde{p}_i = \mathrm{encode}(p_i)
\end{equation}
\begin{equation}
    c_1, c_2, ..., c_k = \mathrm{cluster}(\widetilde{p}_1,\widetilde{p}_2, ..., \widetilde{p}_i)
\end{equation}
\begin{equation}
    d = \{{q_m, r_m, a_m} | \widetilde{p}_m \in c_m, m=1,2,...,k\}
\end{equation}
where $d$ denotes the demonstration set, $c_k$ denotes the $k$-th cluster. Specifically, we use patterns primarily to select demonstrations rather than directly as context for downstream tasks. We utilize the original problem $q_k$ and rationale $r_k$ corresponding to the $p_k$ patterns as the CoT input.

\subsection{Number of Demonstrations}
Since previous methods lack knowledge-based guidance, the choice of $k$ is often based on heuristic values. However, having too many demonstrations does not proportionally enhance the performance \cite{wei2022chain,agarwal2024many}, while too few may fail to adequately capture the task's characteristics. By incorporating reasoning operations, we can use the number of these operations to inform a more reasonable choice for $k$:
\begin{equation}
    k = \lceil \frac{1}{2} \times n \times (1+\mathrm{log}(N)) \rceil
\end{equation}
where $n$ denotes the number of identified operations, and $\lceil \rceil$ represents the ceiling function that rounds up to the nearest integer. This formula empirically takes into account the impact of the number of operation types on the number of demonstrations and further adjusts based on the sample size.

\section{Experiments}
In this section, our objective is to evaluate the effectiveness of our proposed method and answer the following research questions:
\begin{itemize}
    \item \textbf{RQ1}: Does incorporating reasoning patterns enhance the effectiveness of CoT prompting?
    \item \textbf{RQ2}: How do the reasoning patterns influence the outputs of LLMs?
    \item \textbf{RQ3}: Is our method robust and scalable to other models?
\end{itemize}

\begin{table}
    \centering{
    \begin{tabular}{c|c|c}
    \toprule
       Dataset & Samples & Operation Tokens \\
    \midrule
       \multirow{2}{*}{GSM8K} & \multirow{2}{*}{1319} & $+, -, \times, /$\\
       & & `more', `less', `twice', `half' \\
    \midrule
        AQuA & 254 & $+, -, \times, /, \pi, \sqrt{x}, x^n, x^{\circ}, log$ \\
    \midrule
       MultiArith & 600 & \multirow{4}{*}{ $+ , - , \times, /$} \\
       AddSub & 395 & \\
       SingleEq & 508 & \\
       SVAMP & 1000 & \\
    \midrule
       Coin & 500 & `heads up', `tails up' \\
    \midrule
       \multirow{3}{*}{Date} & \multirow{3}{*}{369} & `day', `week',   \\
        & & `month', `year' \\
        & & `yesterday', `tomorrow' \\
    \bottomrule
    \end{tabular}
    }
    \caption{The number of samples and operation tokens.}
    \label{tab:operation}
\end{table}

\subsection{Experimental Setup}
\paragraph{Datasets.} We adopt eight representative datasets for our reasoning tasks: MultiArith \cite{roy-roth-2015-solving}, GSM8K \cite{cobbe2021gsm8k}, AddSub \cite{hosseini-etal-2014-learning}, AQUA-RAT \cite{ling-etal-2017-program}, SingleEq \cite{koncel-kedziorski-etal-2015-parsing}, SVAMP \cite{patel-etal-2021-nlp}, Coin-Flip \cite{wei2022chain}, and BIG-bench Date Understanding \cite{srivastava2023beyond}. They require certain reasoning steps and are commonly used for CoT method comparisons \cite{wei2022chain, kojima2022large, zhang2023automatic,wang2023large,fu2022complexity}.

For tasks MultiArith, AddSub, SingleEq, and SVAMP, we define the set of operation tokens based on a glossary from Wikipedia, as the operations involved are relatively straightforward. For tasks GSM8K and AQUA, we expand the operation token vocabulary manually based on data distribution. For tasks Coin-Flip and BIG-bench Date Understanding, we prompt GPT-4 to generate the corresponding operation tokens. The specific details of the datasets can be found in Table \ref{tab:operation}.

\begin{table*}[]
    \centering{
    \begin{tabular}{@{}clcccccccc}
    \toprule
         \multicolumn{2}{c}{\textbf{LLaMA-2 Model}} & \textbf{MultiArith} & \textbf{GSM8K} & \textbf{AddSub} & \textbf{AQuA} & \textbf{SingleEq} & \textbf{SVAMP} & \textbf{Coin} & \textbf{Date}  \\
         \midrule
         \multirow{6}{*}{7b-chat-hf } & Zero-Shot-CoT & 72.33 & 21.00 & 57.97 & 24.01 & 57.67 & 41.90 & 44.60 & 39.29 \\
         & (+ SC) & \textbf{79.83} & 27.14 & 62.78 & 21.65 & \underline{68.11} & \underline{47.60} & 52.80 & 40.37 \\
         & Random-CoT & 76.16 & 24.41 & \underline{65.59} & 22.44 & 66.14 & 46.59 & 48.00 & 44.44 \\      
         & Auto-CoT & 76.00 & 26.99 & 58.48 & 24.01 & 64.96 & 43.80 & 51.20 & 44.71 \\
         & Auto-CoT-RA & 74.83 & 26.76 & 63.29 & 23.80 & 66.92 & 45.19 & 48.00 & 43.08 \\
         \cmidrule{2-10}
         & Ours & \underline{79.66} & \underline{27.45} & 65.06 & \underline{28.34} & \textbf{71.85} & \textbf{48.50} & \textbf{59.40} & \underline{45.79} \\
         & Ours (Adaptive $k$) & 79.66* & \textbf{28.05} & \textbf{67.08} & \textbf{29.13} & 71.85* & 48.50* & \underline{58.40} & \textbf{46.34} \\
        \midrule
        \midrule
        \multirow{6}{*}{13b-chat-hf} & Zero-Shot-CoT & 77.50 & 34.49 & 60.75 & 15.74 & 69.29 & 49.40 & 47.40 & 46.07 \\  
        & Auto-CoT & \underline{82.16} & 36.77 & 63.03 & 25.19 & \underline{70.67} & \underline{55.50} & 54.20 & 53.93 \\
        & Auto-CoT-RA & 82.16 & 37.04 & 62.08 & \underline{27.74} & 66.14 & 52.10 & \underline{62.80} & 54.47 \\
        \cmidrule{2-10}
        & Ours & \textbf{83.16} & \underline{37.68} & \textbf{65.82} & 26.37 & \textbf{74.80} & \textbf{56.39} & 57.40 & \underline{56.91} \\
        & Ours (Adaptive $k$) & 83.16* & \textbf{38.44} & \underline{64.81} & \textbf{31.49}  & 74.80* & 56.39* & \textbf{67.80} & \textbf{60.97} \\
    \bottomrule
    \end{tabular}
    }
    \caption{Accuracy (\%) on eight reasoning datasets. We present the mean value obtained from five runs. * denotes the situation where $k$ does not change, and results are copied from above. For the Random-CoT method, we report the best result since we are concerned about the potential of CoT. For the self-consistency method, we set the number of paths as 5 \cite{wang2023self}.}
    \label{tab:result}
\end{table*}

\paragraph{Language Models.} To facilitate subsequent interpretability analysis, we select open-source models as our reasoning engine. Specifically, we use models from the LLaMA-2 family due to their foundational logical reasoning capabilities and support for CoT prompting. These models are deployed on our local server, which is equipped with 8 RTX 3090 GPUs, each with 24GB of memory. Due to hardware constraints, we test only the 7B and 13B models. Experiments with larger models or those from other families are discussed in subsequent sections. 

We use the inference functions of these models, and the process does not involve training or fine-tuning. Additionally, we set the hyperparameters with a temperature of 0.4 and top\_p of 0.9 to manage the model's randomness \cite{xu2022systematic}. To maintain consistency with \cite{zhang2023automatic}, we use Sentence-BERT \cite{reimers-2019-sentence-bert} as our encoder and select the \textit{`all-MiniLM-L6-v2'} model for semantic vector representation. This model has also been proven effective in our experiments.

\paragraph{Baselines.} We primarily compare our methods with unsupervised methods including Zero-Shot-CoT \cite{kojima2022large}, Random-CoT, Auto-CoT \cite{zhang2023automatic}, and Self-Consistency \cite{wang2023self}. Building on Auto-CoT, we introduce an additional variant, Auto-CoT-RA, which replaces the original question embedding with the rationale embedding for clustering. The purpose of this modification is to investigate whether this subtle shift can implicitly uncover the underlying patterns in reasoning. Unless otherwise specified, our method uses the same $k$ value as the baseline in experiments. Additionally, we conduct experiments using our method with the adaptive $k$ value that we designed.

\subsection{Main Results (RQ1)}
Table \ref{tab:result} presents the overall performance of various methods on the 7B and 13B models. Since our primary goal is to evaluate whether focusing on diverse patterns provides more benefit to reasoning than semantic information, we are not concerned with identifying which model achieves state-of-the-art performance. Based on these results, we make the following observations:
\begin{itemize}
    \item Overall, our method consistently outperforms the baseline approaches. This stable improvement indicates that by introducing diverse reasoning patterns, we can identify more representative demonstration sets, where each example embodies a different reasoning strategy. Using these diverse examples as context for LLMs can further enhance their ability to solve downstream tasks.
    \item We observe that for arithmetic problems with a limited set of operation tokens, such as MultiArith, AddSub, SingleEq, and SVAMP, our method achieves more significant improvements compared to methods based on semantic information. This suggests that the demonstration sets we construct can effectively cover the majority of reasoning paths, thereby providing comprehensive guidance for LLMs to select appropriate reasoning patterns.
    \item For datasets with a relatively broader action space, like GSM8K and AQuA, the improvements are less significant. This implies that a limited number of examples do not fully capture the diversity of reasoning patterns. However, when we recalculate the number of clusters using adaptive $k$ and expand the demonstration set, we observe additional gains on these two datasets.
    \item Surprisingly, we find that for datasets like Coin and Date, where the operation patterns are not explicitly defined, our method actually lead to greater improvements. We hypothesize that this is because the questions in these datasets are quite similar, making it difficult to distinguish them based on semantic features alone. In contrast, leveraging reasoning patterns allows us to identify representative examples from a different dimension. Additionally, both of these datasets show further performance improvements when using adaptive $k$.
\end{itemize}

\begin{figure}[t]
    \centering
    \includegraphics[width=0.4\textwidth]{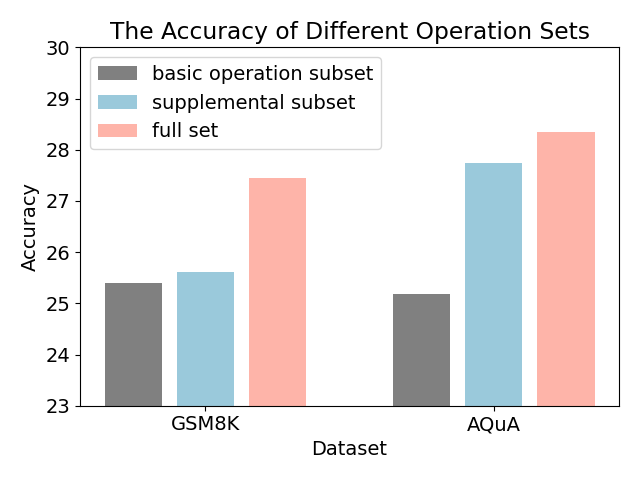}
    \caption{Comparison of different operation sets.}
    \label{fig:subset}
\end{figure}

\begin{table*}[]
        \centering
    \begin{tabular}{p{0.5\linewidth}|p{0.45\linewidth}}
    \toprule
        \textbf{Auto-CoT} & \textbf{Ours} \\
    \midrule
        \textbf{Q:} Mark's father gave him \$85. Mark bought 10 books, each of which cost \$5. How much money does Mark have left? \newline \textbf{A:} Let's think step by step. If Mark has \$85 and he buys 10 books at \$5 each, then \textbf{he will spend a total of \underline{\$50 (10 x \$5)}}. So, after buying the books, \textbf{Mark will have \$35 (\$85 - \$50) left}. The answer is 35. & \textbf{Q:} Tom was at the beach for 5 days and found 7 seashells every day. How many seashells did Tom find during the beach trip? \newline \textbf{A:} Let's think step by step. If Tom found 7 seashells every day, then he found a total of \textbf{7 x 5 = 35} seashells during his entire beach trip. The answer is 35.\\
        \textbf{Q:} Mike has 48 books. Alyssa has 8 times more books than  Mike. How many dozen books does Alyssa have? \newline \textbf{A:} Let's think step by step. If Mike has 48 books, then Alyssa has 8 \textbf{times that number = \underline{8 x 48 = 384}} books. So Alyssa has 384 books. The answer is 384. & \textbf{Q:} Nancy has saved 4900 cents from selling lemonade.  How many dollars does Nancy have? \newline \textbf{A:} Let's think step by step. Nancy saved 4900 cents, which means she saved \textbf{\underline{4900 / 100 = 49}} dollars. The answer is 49.\\
    \midrule
        \textbf{Q:} The value of a sport utility vehicle this year is 16,000 dollars , which is 0.8 of what its value was last year. How much is the value of the vehicle last year? 
         & \textbf{Q:} The value of a sport utility vehicle this year is 16,000 dollars , which is 0.8 of what its value was last year. How much is the value of the vehicle last year? \\
         \textbf{A:} Let's think step by step. The problem tells us that the value of the sport utility vehicle this year is 16,000 dollars, which is 0.8 of what it was last year. So, \textbf{if we multiply the value of the vehicle this year by 0.8, we get the value of the vehicle last year = \underline{16,000 x 0.8 = 12,800} dollars}. The answer is 12,800. & \textbf{A:} Let's think step by step. The problem tells us that the value of the sport utility vehicle this year is \$16,000, which is 0.8 times its value last year. This means that \textbf{the value last year is \underline{\$16,000 / 0.8 = \$20,000}}. The answer is \$20,000. \\
    \bottomrule
    \end{tabular}
    \caption{Case study of Auto-CoT and our method for demonstration selection and downstream inference. The upper questions and answers are demonstrations constructed by two methods, and the lower part contains how LLMs solve the inference task.}
    \label{tab:case}
\end{table*}

\begin{table}[t]
    \centering
    \begin{tabular}{c|c|c|c}
    \toprule
       Dataset & Demos & Incorrect & Error Rate \\
    \midrule
       MultiArith & 8 & 2 & 25.0\% \\
       GSM8K & 8 & 5 & 62.5\% \\
       AddSub & 8 & 3 & 37.5\% \\
       AQuA & 4 & 4 & 100\% \\
       SingleEq & 8 & 2 & 25.0\% \\
       SVAMP & 8 & 6 & 75\% \\
       Coin & 8 & 3 & 37.5\% \\
       Date & 8 & 1 & 12.5\% \\
    \bottomrule
    \end{tabular}
    \caption{The number of demonstrations and their error rate for each dataset.}
    \label{tab:error}
\end{table}

There are several additional observations. For instance, in some cases, Auto-CoT-RA outperforms Auto-CoT, while in others it does not. This suggests that simply shifting from question semantics to rationale semantics does not necessarily narrow the gap between demonstrations and the reasoning task. Deeper reasoning patterns can still be obscured by irrelevant information. Moreover, in certain situations, using a random demonstration set can also surpass Auto-CoT, although this improvement is inconsistent. This indirectly highlights that other factors, such as underlying reasoning patterns, can influence the effectiveness of examples. Our method, in most cases, demonstrates a more stable ability to uncover these factors.

\subsection{Impact of Operation Tokens (RQ1)}
To further assess the impact of reasoning patterns, we conduct additional experiments. Given that GSM8K and AQuA datasets utilize additional operation tokens, we removed some of these tokens to determine their influence. Specifically, we categorize the expanded operation tokens into a basic operation subset, such as $\{+ , - , \times , /\}$, similar to other arithmetic tasks, and the remaining tokens as supplementary subsets. These subsets represent only a portion of the reasoning patterns within these two datasets.

Figure \ref{fig:subset} shows the results of using different subsets on the 7B model. The experimental results demonstrate that using operation subsets as reasoning pattern tokens can degrade overall performance. The primary reason for this is that these subsets do not sufficiently cover the task's logical scope. It leads to a lack of diversity. However, when the full set of operations is utilized, a broader range of scenarios can be activated, allowing the model to better adapt to the task.

\subsection{Case Study (RQ2)}
To gain a deeper understanding of CoT prompting, we perform a case study. Table \ref{tab:case} presents a typical instance analysis. We observe that Auto-CoT, due to its introduction of numerous irrelevant patterns, tends to distort the reasoning results of LLMs. In contrast, our method, which includes a diverse set of reasoning pattern templates, enables the model to generate correct responses.

\begin{figure*}[ht]
  \centering
  \begin{minipage}[b]{0.44\textwidth}
    \includegraphics[width=\textwidth]{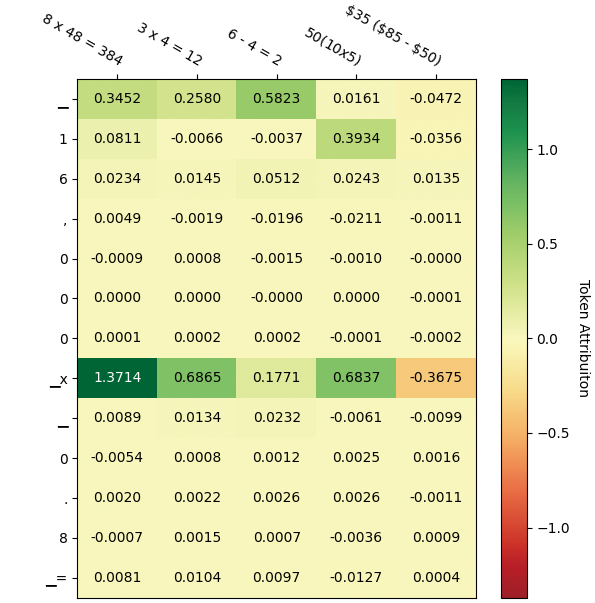}
  \end{minipage}
  \hfill
  \begin{minipage}[b]{0.46\textwidth}
    \includegraphics[width=\textwidth]{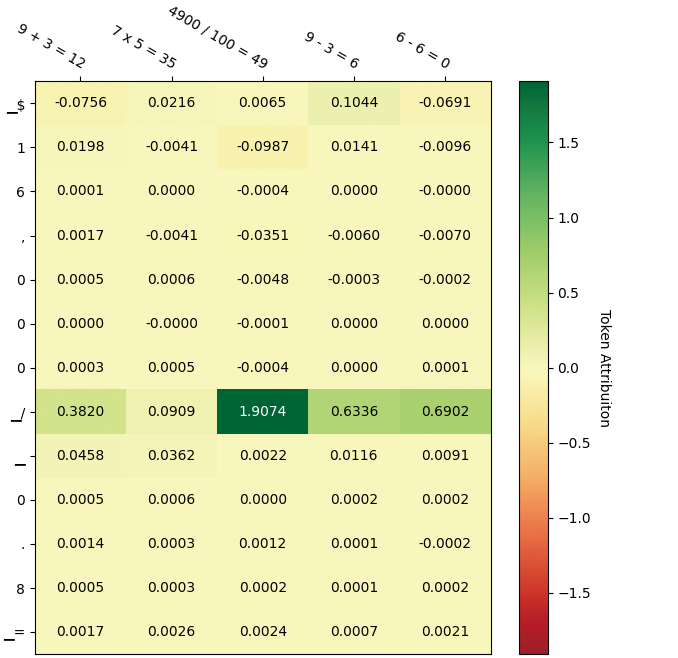}
  \end{minipage}
  \caption{Visualization of token attribution for the case study. The left part stands for the score matrix of patterns from Auto-CoT, and the right part stands for the score matrix from our method. The upper column denotes each individual prompt, and the row denotes the generated token sequence. Higher scores (positive) indicate that the input has a greater impact on the output.}
    \label{fig:attribution}
\end{figure*}

\subsection{Feature Attribution (RQ2)}
Following the previous case study, we seek to understand why different contextual reasoning patterns alter the output of LLMs. Specifically, we employ a perturbation-based feature attribution analysis method \cite{winter2002shapley} to aid in this understanding. Traditional attention-based analysis methods have been criticized for their inability to identify the most significant features \cite{jain2019attention,zhao2024explainability}, which is why we turned to this perturbation-based approach. By masking portions of the input tokens, we recompute the generation probabilities for each output token to assess the input's attribution impact on these output tokens. We use Captum \cite{miglani2023using} to achieve this visualization. Figure \ref{fig:attribution} presents the attribution analysis matrix for the case study. According to the visualization results, we find that when a particular pattern is overly dense in the examples, the model tends to activate related knowledge, which can lead to biased reasoning processes. Conversely, when these patterns are more diverse, the model is more likely to activate the correct reasoning pathways. Our method, by enhancing the diversity of patterns in the demonstrations, effectively reduces the distance to the reasoning task objectives.

\begin{table}
    \centering{
    \begin{tabular}{@{}clcccc}
    \toprule
        \multicolumn{2}{c}{\textbf{Model}} & \textbf{AddSub} & \textbf{AQuA} & \textbf{SingleEq} \\
         \midrule
         \multirow{3}{*}{GPT-3.5} & Zero-Shot & 83.29 & 59.44 & 90.55 \\
         & Auto-CoT & 81.26 & 58.66 & 91.53 \\
         & Ours & \textbf{83.54} & \textbf{62.38} & \textbf{93.11} \\
    \midrule
         \multirow{3}{*}{Qwen} & Zero-Shot & 54.93 & \textbf{35.03} & 69.07 \\
         & Auto-CoT & 62.53 & 30.31 & 80.31 \\
         & Ours & \textbf{67.59} & 33.46 & \textbf{82.08} \\     
    \bottomrule
    \end{tabular}
    }
    \caption{Result of GPT-3.5-turbo-0125 and Qwen-7b-chat model on different datasets.}
    \label{tab:gpt}
\end{table}

\subsection{Error Robustness (RQ3)}
It is worth mentioning that we do not impose supervision on the labels of the demonstrations. Therefore, we proceed to count the number of incorrect instances within the selected set, as shown in Table \ref{tab:error}. It is intriguing to notice that the majority of our provided demonstrations are imperfect, with AQuA even exhibiting a 100\% error rate. This phenomenon suggests that LLMs struggle to discern incorrect instances from correct ones. Instead, they learn from how the example approaches problem-solving, which we refer to as `pattern'. Our method encourages LLMs to follow the most probable reasoning chain towards the final answer and thus leads to a significant improvement.

\subsection{Results on Other Models (RQ3)}
To determine whether our method is applicable to different models, we test it on various LLM branches. Specifically, we select the GPT series to represent larger and more advanced models, and Qwen to represent multilingual models. For the sake of hardware resources and budget constraints, we experiment with the GPT-3.5-turbo and Qwen-7B models. Table \ref{tab:gpt} presents the performance of several methods on these models. Notably, the experiments show that Auto-CoT, in some cases, underperforms compared to direct answering on these models. We attribute this to the inherent noise in semantics-based methods. Our approach mitigates this noise, resulting in more consistent performance improvements.

\section{Conclusion}
This paper aims to address the noise issue inherent in unsupervised semantic-based CoT methods and proposes a reasoning pattern-based approach for CoT demonstration selection. Our method explicitly enhances the interpretability of reasoning processes and illustrates how LLMs can be guided toward generating accurate answers. Extensive experiments validate the effectiveness, robustness, and compatibility of our approach.

\section{Acknowledgements}
This work was supported by the National Key R\&D Program of China (Grant No.2023ZD0120400), Beijing Natural Science Foundation (L247028), National Natural Science Foundation of China (No. 62276260, 62076235), Beijing Municipal Science and Technology Project (Z231100007423004). We sincerely thank all reviewers and ACs for their insightful comments, time and efforts.

\bibliography{aaai25}

\end{document}